\documentclass[preprint]{article}

\usepackage{neurips_2025}

\usepackage[utf8]{inputenc} 
\usepackage[T1]{fontenc}    
\usepackage{hyperref}       
\usepackage{url}            
\usepackage{booktabs}       
\usepackage{amsfonts}       
\usepackage{nicefrac}       
\usepackage{microtype}      

\newtheorem{definition}{Definition}
\usepackage{graphicx}
\usepackage{subcaption}
\usepackage{amsmath}
\usepackage{mathrsfs}
\usepackage[ruled,vlined,linesnumbered]{algorithm2e}
\usepackage{array}
\usepackage[table]{xcolor}
\usepackage{float}
\usepackage{wrapfig}
\usepackage{setspace}
\usepackage{colortbl}

\title{Sampling Imbalanced Data with Multi-objective Bilevel Optimization}

\author{%
  Karen ~Medlin\\
  Department of Mathematics\\
  University of North Carolina at Chapel Hill\\
   \And
    Sven Leyffer \\
  Mathematics and Computer Sciences Division \\
  Argonne National Laboratory \\
   \AND
   Krishnan Raghavan \\
  Mathematics and Computer Sciences Division \\
  Argonne National Laboratory \\
}

\UseMicrotypeSet[protrusion]{basicmath} 
\microtypesetup{protrusion=false} 
\begin{document}

\maketitle

\begin{abstract}
Two-class classification problems are often characterized by an imbalance between the number of majority and minority datapoints resulting in poor classification of the minority class in particular. Traditional approaches, such as reweighting the loss function or naïve resampling, risk overfitting and subsequently fail to improve classification because they do not consider the diversity between majority and minority datasets. Such consideration is infeasible because there is no metric that can measure the impact of imbalance on the model. To obviate these challenges,  we make two key contributions. First, we introduce MOODS~(Multi-Objective Optimization for Data Sampling), a novel multi-objective bilevel optimization framework that guides both synthetic oversampling and majority undersampling. Second, we introduce a validation metric-- `$\epsilon/ \delta$ non-overlapping diversification metric' that quantifies the goodness of a sampling method towards model performance. With this metric we experimentally demonstrate state-of-the-art performance with improvement in diversity driving a $1-15 \%$ increase in $F1$ scores.
\end{abstract}

\section{Introduction}\label{section: Intro}
Imbalanced datasets pose challenges across many domains, as models trained on such data often produce unreliable predictions. In binary classification, this issue arises when one class (majority) vastly outnumbers the other (minority), leading to models biased toward the majority. Examples include disease detection \cite{bria_addressing_2020,johnson2019survey, khan2018cost-sensitive}, identifying chemicals as either harmful or beneficial \cite{bae2021effective, korkmaz2020chemicaljournal}, predicting natural disasters \cite{kumar2021classification}, and detecting fraudulent transactions\cite{singh2022credit, makki2019experimental}, among others \cite{kanika2020survey, haixiang2017learning}.

Common solutions address imbalance by reweighting the loss function~\cite{dang_inverse_2024, pmlr-v162-guo22e} or rebalancing the training data~\cite{strelcenia_survey_2023, das_supervised_2022, kovacs_empirical_2019}. Reweighting loss often causes overfitting to the minority class, while rebalancing data—via undersampling or duplicating points—only adjusts sample counts. This, too, can create an issue of overfitting especially when the training data under-represents some region of the actual data distribution. In this scenario, overfitting occurs as adjusting sample counts alone does not take into account regional under-representations of a dataset. For example, generating more upright zeros in MNIST \cite{deng_MNIST_2012} does not prepare a model for rotated versions of zeros. Naïve sampling methods therefore do not capture data diversity which is essential for better model performance. A better strategy is to generate synthetic samples that reflect missing characteristics — e.g., rotated versions of a digit—to improve robustness. Therefore, we argue that new samples must (i) enhance minority diversity and (ii) avoid overlapping with majority data.

\begin{wrapfigure}[16]{r}{0.6\textwidth}
    \centering
    \vspace{-10pt}
    \includegraphics[width = 0.58\textwidth]{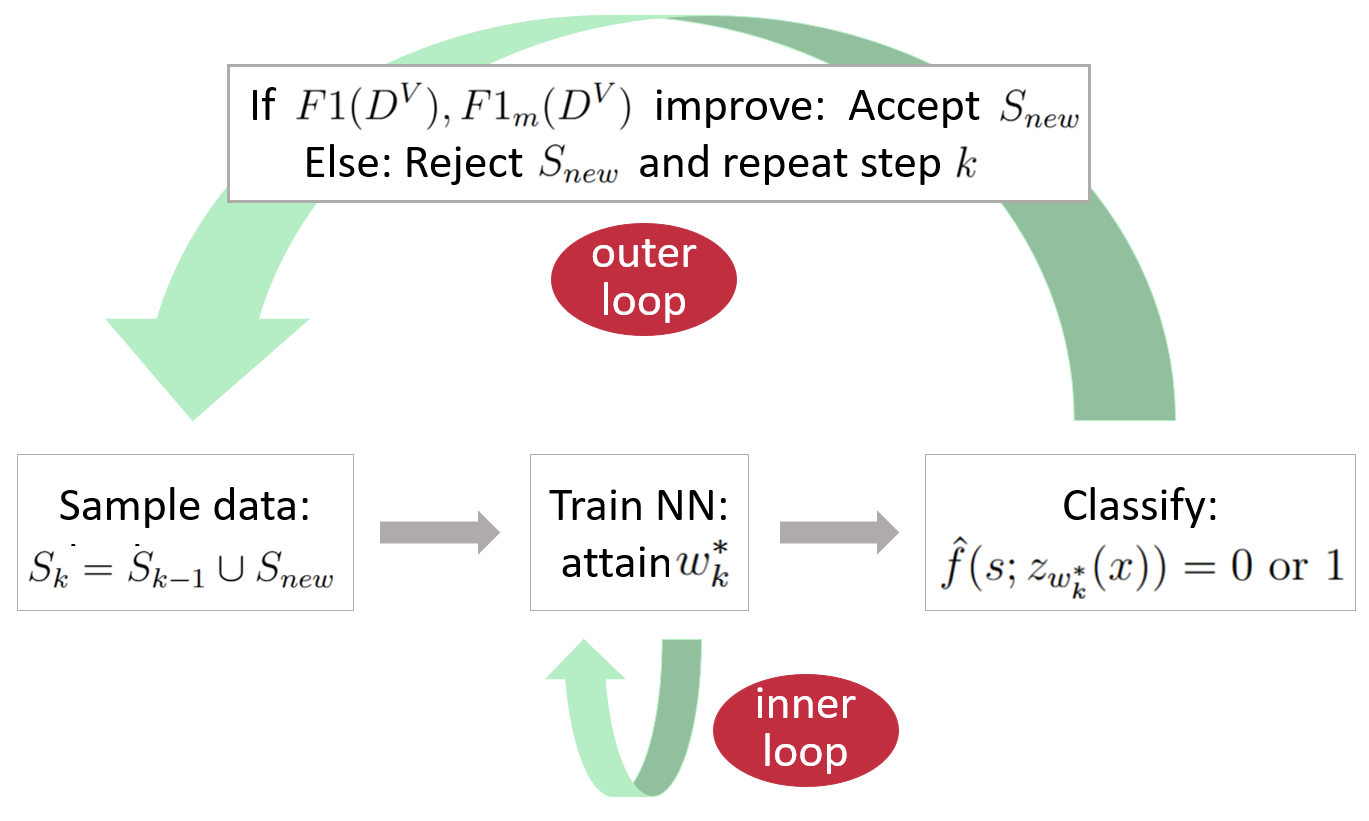}
    \caption{MOODS accepts/rejects samples of training data while iterating over the inner- and outer-loop for each step $k$.}
    \label{fig:moodsGraphic}
    \vspace{-5pt}
\end{wrapfigure}

To meet these requirements, we leverage SVM-SMOTE \cite{nguyen_borderline_2011} to generate synthetic data while winnowing both majority data and synthetic minority data with a bilevel optimization framework. We designed a two-step method, MOODS: Multi-Objective Optimization for Data Sampling~(see Fig. \ref{fig:moodsGraphic}/Prob. \ref{bilevelProblem}), where an upper-level/outer loop generates samples with increased diversity and decreased overlap by maximizing overall and minority $F1$ scores (Def. \ref{def: F1score}) while the lower-level/inner loop learns model weights minimizing loss with these generated samples.

\begin{wrapfigure}{1}{0.6\textwidth}
    \centering
    \vspace{-35pt}
    \begin{subfigure}[b]{0.55\textwidth}
        \includegraphics[width = \textwidth]{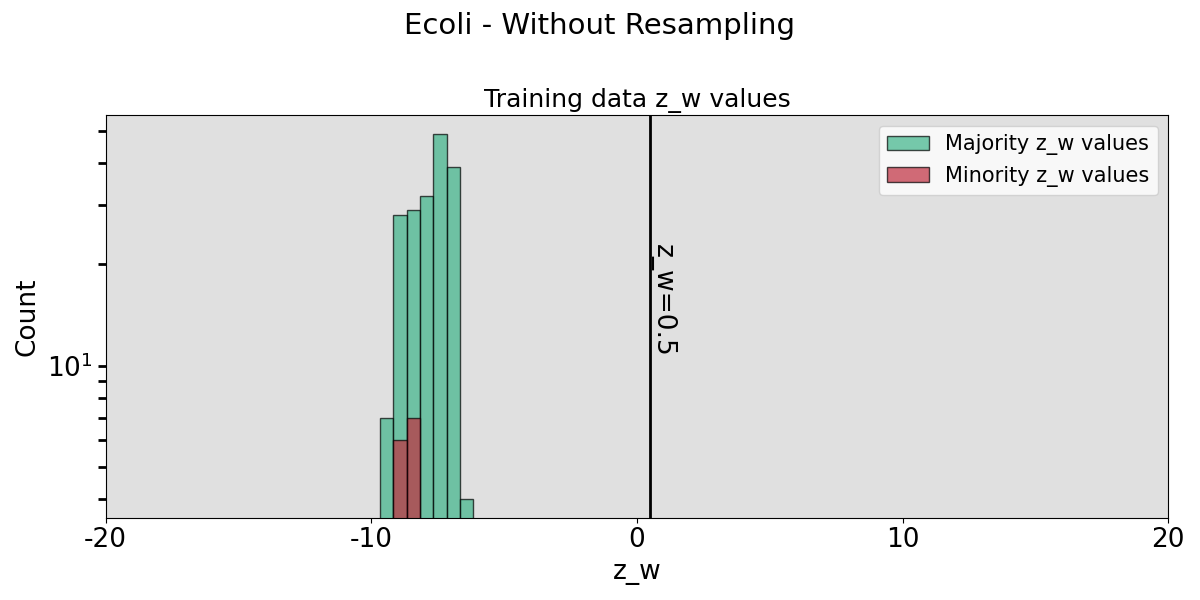}
    \end{subfigure}

    \vspace{0.75em}
    
   \begin{subfigure}[b]{.55\textwidth}
        \includegraphics[width = \textwidth]{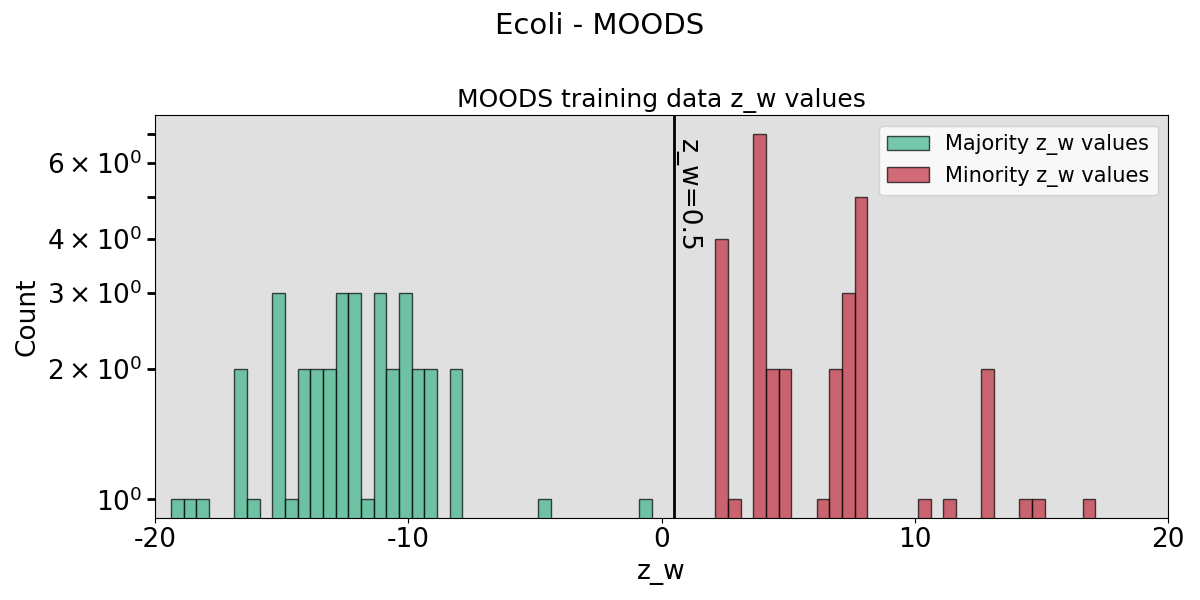}
    \end{subfigure}
    \caption{The histograms show $z_{w^*}$ model outputs from imbalanced Ecoli data (top) and MOODS resampled data (bottom). The model architecture in both scenarios is the same. When considering $z_{w^*} \in \mathbb{R}$ outputs as proxy measurements for feature data $x \in \mathbb{R}^n$ inputs (Sec.~\ref{novelMetricSection}), we see that the MOODS sampling technique supports the separation of training data into balanced and diverse minority (red; right) and majority (green; left) data without overlap. This results a strong $0.95$ $F1$ score.}
    \vspace{-5pt}
    \label{fig:zEcoli}
\end{wrapfigure}

To measure the diversity and overlap of the generated samples and, by extension, the effectiveness of learning, we introduce a novel `$\epsilon/ \delta$ non-overlapping diversification metric.' For instance as indicated in Fig.~\ref{fig:zEcoli}, our MOODS sampling approach resulted in decreased minority $z_{w^*}$ overlap (Def. \ref{def:minorityOverlap}) by $100\%$ and increased $z_{w^*}$ diversity (Def. \ref{Def: deltaVariance}) by over two orders of magnitude for the Ecoli dataset, which corresponds to a strong $F1$ score of $0.95$. We demonstrate that improvement in diversity and overlap as indicated by our metric coincides with a $1\%$ to $15\%$ improvement in $F1$ scores with a comprehensive simulation study including seven imbalanced datasets and seven state-of-the-art imbalanced classification methods.

\subsection{Related works}
Classification methods for imbalanced data fall into two main categories: those that rebalance training data and those that rebalance neural network models. We focus this section on surveying \textit{data} rebalancing methods as this aligns with our approach. Comprehensive surveys on imbalanced data classification including model-based approaches can be found in \cite{strelcenia_survey_2023, das_supervised_2022, kovacs_empirical_2019}.

SMOTE and its variants, such as SMOTE+TomekLinks \cite{batista_study_2004}, Borderline-SMOTE \cite{hutchison_borderline-smote_2005}, and SVM-SMOTE \cite{nguyen_borderline_2011}, have dominated imbalanced data classification for over two decades. However, their heavy reliance on synthetic data often leads to class overlap. Recent methods like IMWMOTE \cite{wang_imwmote_2024}, Reduced-Noise SMOTE \cite{badawy_rn-smote_2022}, and Radius-SMOTE \cite{Pradipta_RadiusSMOTE} aim to mitigate overlap by refining synthetic data apart from the majority class. Still, oversampling methods suffer from overfitting. Techniques like Diversity-based Selection \cite{yang_diversity-based_2023}, BalanceMix \cite{song_toward_2023}, and RIO \cite{chang_image-level_2021} address this by promoting diverse synthetic data.

GANs have emerged as a popular tool to tackle overlap and lack of diversity and in SMOTE-based approaches particularly. For instance, CTGAN-MOS \cite{Majeed_CTgan} and SVDD \cite{ding_leveraging_2024} reduce noise, LEGAN \cite{ding_legan_2024} enhances feature diversity, and HD-GNNS \cite{liu_hybrid_2024} and ImbGAN \cite{yun_learning_2024} decrease overlap. Hybrid approaches, such as SMOTified-GAN \cite{sharma_smotified-gan_2022}, combine SMOTE-generated samples with GANs to refine synthetic data. GAN-based Oversampling (GBO) \cite{ahsan_enhancing_2024} and SVM-SMOTE-GAN (SSG) \cite{ahsan_enhancing_2024} further reduce overlap by interpolating between minority class samples. A recent survey \cite{cheah_enhancing_2023} of these SMOTE-GAN approaches found that they perform well for detecting financial fraud. Despite these advances, overlap and lack of diversity continue to challenge classification, and GANs suffer from training instability and mode collapse in which the generator produces only the same output. Additionally, none of these approaches introduce a reliable way of measuring diversity.

To avoid the pitfalls of using a GAN, our proposed approach uses a multi-objective bilevel optimization framework to sort through samples of training data and construct an optimal training set. While bilevel optimization has been previously employed in the context of imbalanced data classification \cite{rosales-perez_handling_2023, chabbouh_imbalanced_2023, hammami_feature_2020}, to the best of our knowledge the only other work that explicitly applies bilevel optimization to rebalance training data is Majority Undersampling with Bilevel Optimiation (MUBO) \cite{medlin_bilevel_2024}. Unlike MUBO: (a) we use a multi-objective formulation in the upper- level problem whereby both the sample $F1$ and minority $F1$ scores are maximized~(MUBO minimizes majority loss alone); (b) the objective of our algorithm’s outer loop is to get to a single Pareto optimal point (see Fig. \ref{fig:Pareto}); and (c) we use a small amount of synthetic data while MUBO uses none. These differences result in MOODS's stronger performance on a wider assortment of data (Table \ref{fig:moodsResults}).  

Our method builds on SVM-SMOTE, which we use to generate synthetic data, as does SSG to which we compare MOODS. SVM-SMOTE generates synthetic points to bolster the minority side of the decision boundary. Our bilevel optimization tests whether these synthetic points improve classification, selecting for diverse feature data and refining SVM-SMOTE’s ability to reduce overlap. Sec. \ref{evaluationSection} compares our results with SVM-SMOTE, SMOTified-GAN, GBO, SSG, and MUBO. Like our approach, SMOTified-GAN, GBO, SSG, and MUBO all try to improve binary classification with sampling techniques that address issues of overlap and lack of diversity.

To clarify, established bilevel optimization methods resemble our approach in how their lower-level problems minimize loss over model parameters. For example, AutoBalance \cite{li_autobalance_2022} uses a bilevel framework to classify imbalanced data by optimizing loss over parameters and hyperparameters. Similarly, network architecture search methods define lower-level objectives to minimize model loss \cite{zhou_theory-inspired_2020, liu_darts_2019, luketina_scalable_2016}. Additionally, \cite{SODA20111801} uses multi-objective optimization for imbalanced classification to find optimal classification models. Our method significantly differs from these methods as demonstrated in its upper-level problem formulated to find optimal training data (see Fig. \ref{fig:moodsGraphic} and Sec. \ref{bilevelSection}). 

\section{Background}\label{background}
This section provides background material for our bilevel optimization problem and classification algorithm in Sec. \ref{bilevelSection}; our `$\epsilon/ \delta$ non-overlapping diversification' metric in Sec. \ref{novelMetricSection}; and experimental results in Sec. \ref{evaluationSection}. We begin with a set of imbalanced data $D = D_m \cup D_M$ with elements $s=(x,y) \in D,$ feature vectors $x \in \mathbb{R}^n$, and labels $y \in \{0,1\}$. We use subscripts $m$ for minority and $M$ for majority sets, and denote the number of elements in a set with $|\cdot|$. Dataset $D$ is partitioned into (disjoint) sets for training, validation, and testing - $D = D^{Tr} \cup D^{V} \cup D^{Te}$. Note that $D^{Tr} \cap D^{V} = D^{V} \cap D^{Te} = D^{Te} \cap D^{Tr} = \emptyset$; $|D^{Tr}| > |D^{Te}|$; and $|D^V| \sim |D^{Te}|$.

In our algorithm, we iteratively update a training set, $S^{Tr}$, which contains all of a dataset's minority data plus synthetic minority data; i.e., $S^{Tr} = S_m^{Tr} \cup S_M^{Tr} =  \underbrace{S_m^{Sy} \cup D^{Tr}_m}_{S_m^{Tr}} \cup  \underbrace{S^{Tr}_M}_{\subseteq D^{Tr}_M}$ where  $S_m^{Sy}$ is synthetic minority data. In contrast to our dynamic training set, $S^{Tr}$, our testing and validation sets, $D^{Te}$ and $D^{V}$, contain only data from the originial dataset $D$. Since we are focused on the two class case, we use the loss function commonly referred to as ``binary cross entropy'' \cite{bishop1996} to train our model with samples of minority and majority training data $S=S_m \cup S_M$; $S \subseteq S^{Tr}$.

\begin{definition}
Let $\mathscr{L}(S;z_w(x))  $ denote \textbf{sample loss} as follows 
\begin{footnotesize} 
\begin{align*}
    \mathscr{L}(S; z_w(x)) 
    &= \underbrace{\frac{1}{|S|}\sum_{s \in S_m}-\log(\frac{e^{z_w(x)}}{ e^{z_w(x)} + e^{1-z_w(x)}})}_{J(S_m; z_w(x))}\quad + \underbrace{\frac{1}{|S|}\sum_{s \in S_M}-\log(\frac{e^{1-z_w(x)}}{ e^{z_w(x)} + e^{1-z_w(x)}})}_{J(S_M; z_w(x))},\\
    \end{align*}
\end{footnotesize}%
where $z_w(x) \in \mathbb{R}$ is the output of a nonlinear neural network dependent on model parameters $w$; and $J(S_m; z_w(x))$ and $J(S_M ;z_w(x))$ denote \textbf{minority loss} and \textbf{majority loss}, respectively.
\end{definition}

Sample loss is optimized over model parameters $w$ and appears in the lower-level optimization problem 
\ref{bilevelProblem}.
Unlike loss, which depends on the class to which a datapoint $s$ actually belongs, classification depends on which of the two classes $s$ is more likely to belong. 

\begin{definition}\label{pointClassification}
Let $\hat{f}(s;z_{w^*}(x))$ denote \textbf{classification of a datapoint $s\in S$} as follows 

\begin{footnotesize} 
\begin{align*}
\hat{f}(s;z_{w^*}(x))
&= \begin{cases} 
   0,  \text{ if } \frac{e^{1-z_{w^*}(x)}}{ e^{z_{w^*}(x)} + e^{1-z_{w^*}(x)}} \geq \frac{e^{z_{w^*}(x)}}{ e^{z_{w^*}(x)} + e^{1-z_{w^*}(x)}} \\ 
   1,  \text{ else},
\end{cases}\\
\end{align*}
\end{footnotesize} 
where $w^* = \underset{w}{\text{argmin}} \quad \mathscr{L}(S;z_w(x))$ denotes the \textbf{optimal parameters for a given training sample $S$.}
\end{definition}

Classification is correct only if $\hat{f}(s;z_{w^*}(x))=y$. It is important to note that by Def. \ref{pointClassification}, $\hat{f}(s;z_{w^*}(x)) = 0 \iff z_{w^*}(x) \leq \frac{1}{2}$. We rely on this idea when defining minority overlap (Def. \ref{def:minorityOverlap}) as part of our `$\epsilon/ \delta$ non-overlapping diversification' metric in Sec. \ref{novelMetricSection}. 

Given majority data's dominance, accuracy skews model performance towards majority data, resulting in high scores regardless of minority performance. Unlike accuracy, the average $F1$ score considers minority and majority performance without skew and is often the preferred metric for imbalanced data classification \cite{jeni_facing_2013}. 

\begin{definition}\label{def: F1score}
Let $F1(D^{V};z_{w^*}) \in [0,1]$ denote the \textbf{average $F1$ score of validation data} as follows
\end{definition}
\begin{footnotesize}
\begin{align*}
    F1(D^{V};z_{w^*}) &= \frac{1}{2}\underbrace{\left(\frac{2T_p(D^{V})}{2T_p(D^{V}) + F_p(D^{V}) + F_n(D^{V})}\right)}_{F1_m(D^{V};z_{w^*})} +
    &\frac{1}{2}\underbrace{\left(\frac{2T_n(D^{V})}{2T_n(D^{V}) + F_n(D^{V}) + F_p(D^{V})}\right)}_{F1_M(D^{V};z_{w^*})},\\
\end{align*}
\end{footnotesize}%
where $F1_m(D^{V};z_{w^*})$ and $F1_M(D^{V};z_{w^*})$ denote the \textbf{minority} and \textbf{majority $F1$ score of validation data}, respectively,
and where $T_p(D^{V}) = \sum_{s\in D^V_m} \hat{f}(s;z_{w^*}(x)); \quad F_n(D^{V}) = \sum_{s\in D^V_m} (1-\hat{f}(s;z_{w^*}(x))) = |D^V_m|-T_p(D^{V});$ $F_p(D^{V}) = \sum_{s \in D^V_M} \hat{f}(s;z_{w^*}(x));\quad\text { and  } T_n(D^{V}) = \sum_{s\in D^V_M} (1-\hat{f}(s;z_{w^*}(x))) = |D^V_M| - F_p(D^{V}).$ To note, $F1(D^V;z_{w^*})$ and $F1_m(D^V;z_{w^*})$ appear in the upper-level problem \ref{bilevelProblem} and are used to optimize training data resulting in the \textbf{final sample of training data $\mathsf{S}$}. The \textbf{$F1$ scores of testing data} - $F1(D^{Te};z_{w^*})$, $F1_m(D^{Te};z_{w^*})$, $F1_M(D^{Te};z_{w^*})$ - are used in Sec. \ref{evaluationSection} to assess performance.

\section{Sampling with Bilevel Optimization}
\label{bilevelSection}
Our bilevel problem is formulated as follows:
\begin{align}
    &\underset{S \subset S^{Tr}}{\text{min}} \quad \left[(1-F1_m(D^{V}; z_{w^*}(x))),(1 - F1(D^{V}; z_{w^*}(x)))\right]\nonumber \\
    &\text{ s.t. } \quad w^* = \underset{w}{\text{argmin}} \quad \mathscr{L}(S;z_w(x)).\label{bilevelProblem}
\end{align}  

Alg. \ref{MOODSalgo} aims to solve this multi-objective bilevel optimization problem Prob. \ref{bilevelProblem} and, thereby, construct an optimal training set. Note, the upper-level problem maximizes both minority and average $F1$ scores over samples of training data. Optimizing this upper-level objective corresponds to the outer loop in Alg. \ref{MOODSalgo}, which is represented by the upper green arrow in Fig. \ref{fig:moodsGraphic} indicating resampling the training set. The lower-level objective in \ref{bilevelProblem} is optimized in the inner loop in Alg. \ref{MOODSalgo}, which is represented by the lower green arrow in Fig. \ref{fig:moodsGraphic}. We note that the two loops do not alternate. At every iteration of the outer loop, we completely solve the training problem (inner loop). Bilevel optimization is also referred to as leader-follower games and is a sequential decision model where the follower (lower level) chooses optimal weights given the leader's (upper level) choice of sample set $S$. In this sense, the leader can anticipate the follower's decision and takes the optimal weights into account when choosing $S$. A multi-objective bilevel optimization framework lends itself well to other classification methods and learning more generally. The upper-level problem can be swapped out for achieving particular learning objectives.

\subsection{MOODS: Multi-Objective Optimization for Data Sampling}
\label{algo}

\begin{wrapfigure}{R}{0.55\textwidth}
\vspace{-1em}  
\hspace*{1em} 
\begin{minipage}{0.95\linewidth}

\setstretch{0.95}  

\begin{algorithm}[H]
\caption{\textbf{MOODS}: Multi-Objective Optimization for Data Sampling}
\label{MOODSalgo}
\KwIn{Training data $S^{Tr} = S_m^{Tr} \cup S_M^{Tr}$, validation data $D^V$}
\KwOut{Final training set $\mathsf{S} = S_k$}

$M_0,M_1 \gets \lceil \frac{1}{2}|S_m| \rceil$; $S_{M_0} \gets \text{randint}(S_M^{Tr}, M_0)$; 
$S_0 \gets S_m^{Tr} \cup S_{M_0}$; 
$S_M^{Tr} \gets S_M^{Tr} \setminus S_{M_0}$\;
Initialize $p(s) \gets \frac{1}{|S_M^{Tr}|}$ for all $s \in S_M^{Tr}$; $k \gets 1$\;

\While{$k <$ MaxIter}{
    Normalize $p(s)$ over $S_M^{Tr}$\;
    $S_{M_k} \gets \text{randint}(S_M^{Tr}, M_k)$\;
    $\hat{S}_{m_k} \gets \text{svm-smote}(S_{k-1} \cup S_{M_k})$\;
    $S_k \gets S_{k-1} \cup S_{M_k} \cup \hat{S}_{m_k}$\;

    \tcc{Inner loop: train model and compute validation metrics}
    $epochs \gets 0$\;
    \While{$\lVert \nabla \mathscr{L} \rVert >$ GradTol \textbf{and} $epochs <$ MaxEpochs}{
        Optimize $\mathscr{L}(S_k; z_w)$\;
        $epochs \gets epochs + 1$\;
    }

    $z_{w_k^*} \gets$ trained model\;
    $F1_m^{(k)} \gets F1_{minority}(D^V; z_{w_k^*})$\;
    $F1^{(k)} \gets F1_{overall}(D^V; z_{w_k^*})$\;
    \tcc{Outer loop: check improvement and accept or reject}
    \If{$(1 - F1^{(k)}) < (1 - F1^{(k-1)})$ \textbf{and} $(1 - F1_m^{(k)}) < (1 - F1_m^{(k-1)})$}{
        Accept step: keep $S_k$\;
        $S_M^{Tr} \gets S_M^{Tr} \setminus S_{M_k}$\;
        $M_k \gets M_k + 1$\;
    }
    \Else{
        Reject step: revert $S_k \gets S_{k-1}$\;
        Halve $p(s)$ for all $s \in \hat{S}_{m_k}$\;
        $M_k \gets M_k - 1$; $k \gets k - 1$\;
    }

    $k \gets k + 1$\;
}
\Return $S_k$\;
\end{algorithm}
\end{minipage}
\vspace{-2em}
\end{wrapfigure}

As depicted in Fig.\ref{fig:moodsGraphic}, the lower- and upper-level problems of the bilevel optimization Prob. \ref{bilevelProblem} work in tandem and correspond to the inner and outer loops of the MOODS Alg. \ref{MOODSalgo}. By minimizing $(1-F1_m(D^{V}; z_{w^*}(x)))$ and $(1 - F1(D^{V}; z_{w^*}(x)))$ over samples of training data, the upper-level problem of the outer loop selects only those samples of training data that improve both average and minority $F1$ scores. To prevent overfitting, validation data $D^V$ that is disjoint from  training data $S^{Tr}$ is used to compute $F1$ scores of the upper-level problem. 

Our algorithm begins with roughly the same number of minority and majority datapoints (line 1 in Alg. \ref{MOODSalgo}) and bit-by-bit increases or decreases both numbers of datapoints by $\pm 1$ as step $k$ samples are either accepted or rejected (lines 6, 18, 22). Computational costs come primarily from the inner loop, when sample loss is minimized over parameters (line 10) as is common for neural network learning. Each step $k$ incurs this expense so as to compute whether or not the new training sample improves $F1$ scores. 

Experimentally we confirmed that our multi-objective bilevel optmization approach balances both of its objectives, identifying a single Pareto optimal point (Fig. \ref{fig:Pareto}). The particular point that we identify is based purely on the progress/sampling of the method, which is somewhat unsatisfactory. We did experiment with an approach that combined both objectives, and just required standard descent. Despite strong empirical findings, we did not find a Pareto front so have not yet confirmed convergence nor that the objective is being achieved. To verify our findings, we developed the $\epsilon/ \delta$ non-overlapping diversification metric (Sec. \ref{novelMetricSection}) that measures a sampling method's ability to deliver a diverse and non-overlapping training set.

\section{A Novel Metric for Sampling Techniques and Their Resampled Data}\label{novelMetricSection}
While \textit{classification performance} of models trained with different sets of resampled training data can be measure with $F1$ scores, how can \textit{sampling performance} of a sampling technique itself be measured? It is generally agreed that superior sampling techniques construct diverse and non-overlapping training sets. As such, we propose a novel metric to measure a sampling method's ability to resample imbalanced data into non-overlapping and diverse training sets. We use this metric to verify the optimality of $\mathsf{S}$, MOODS's final resampled training set (Sec. \ref{evaluationSection}).  

\subsection{Model Outputs as 1-D Analog to Model Inputs $x \in \mathbb{R}^n$}\label{oneD}

To define overlap and diversity, we first choose  a model's output $z_{w^*}(x) \in \mathbb{R}$  (Def. \ref{def: F1score}) to be 1-D analog to its model input $x \in \mathbb{R}^n$. We do this because, for one, diversity metrics such as variance are more illuminating in $\mathbb{R}$ than they are in $\mathbb{R}^n$. Since $z_{w^*}(x)$ model outputs depend on both training data \textit{and} model architecture, to make model outputs analogous to training data inputs alone, we keep the model architecture constant (Sec. \ref{section: setUp}). Let $z_{w^*}=z_{w^*}(x)$

\subsection{Defining $\epsilon/ \delta $ Non-overlapping Diversification} 
To show MOODS's ability to successfully resample imbalanced data into diverse and non-overlapping sets, we construct the \textbf{$\epsilon/ \delta$ non-overlapping diversification metric}. The metric compares model outputs of the initial, imbalanced training data, $D^{Tr}$, with model outputs of MOODS's final resampled training set, $\mathsf{S}$. The model has the same architecture for both datasets. The metric can be applied generically to any two sets of training data, $S_1$ and $S_2$, by substituting $S_1$ for $D^{Tr}$ and  $S_2$ for $\mathsf{S}$. 

Recall $S=S_m \cup S_M$ is a sample of training data. To define overlap, we need a minority/majority decision boundary. Since for all $x \in S$, $e^{z_{w^*}}= e^{1-z_{w^*}} \iff z_{w^*} = \frac{1}{2}$ (Def. {\ref{pointClassification}), $z_{w^*} =\frac{1}{2}$ is a clear choice for the minority/majority decision boundary.  We use $\widehat{S_m} \subset S_m$ to denote a set of \textbf{overlapping minority points} whose $z_{w^*}$ values are on the majority (incorrect) side of $z_{w^*} = \frac{1}{2}$ where $\widehat{S_m} := \bigl\{s = (x,1) \mid z_{w^*} \leq \frac{1}{2}\bigr \}$. To show our method's ability to improve minority overlap, we measure the decrease in minority $z_{w^*}$ overlap from the initial training set, $D^{Tr}$, to the final resampled set, $\mathsf{S}$ with the following definition.  
\begin{definition}\label{def:minorityOverlap}
    Let $\Delta_{D^{Tr}, \mathsf{S}}(\kappa_m) \in [0,100]$ denote  \textbf{decrease in minority $z_{w^*}$ overlap}:
\begin{align*}
    \Delta_{D^{Tr}, \mathsf{S}}(\kappa_m) &:= \left(\frac{|\widehat{D^{Tr}_m}|}{|D^{Tr}_m|} - \frac{|\widehat{\mathsf{S}_m}|}{|\mathsf{S}_m|}\right) \times 100\%.
\end{align*}
\end{definition}

For example in Fig.\ref{fig:zEcoli}, Ecoli's improvement in minority $z_{w^*}$ overlap, $\Delta_{D^{Tr}, \mathsf{S}}(\kappa_m)$, is $1 \times 100\% = 100\%$. To compute this, we see that $\frac{|\widehat{D^{Tr}_m}|}{|D^{Tr}_m|}=1$ as all of Ecoli's unsampled minority outputs (red columns; top) are on the wrong side of the $z_{w^*} = \frac{1}{2}$ boundary. On the other hand, $\frac{|\widehat{\mathsf{S}_m}|}{|\mathsf{S}_m|}=0$ because all of Ecoli's MOODS resampled minority outputs (red columns; bottom) are on the correct side. 

To express a sampling method's ability to improve feature diversity, we use $s^2(S)$ to denote \textbf{$z_{w^*}$ variance} for all $x \in S$. We measure variance increase from $D^{Tr}$ to $\mathsf{S}$. Due to datasets having wide ranging variances, we measure increases by orders of magnitude. We use $\Delta O_{D^{Tr},\mathsf{S}}(s^2_m) \in \mathbb{N}$ to denote \textbf{increase in minority $z_{w^*}$ variance} where $\Delta O_{D^{Tr}, \mathsf{S}}(s^2_m) = \left\lfloor \log_{10}\left(\frac{s^2(\mathsf{S}_m)}{s^2(\mathsf{D}^{Tr}_m)}\right) \right\rfloor$, and $\Delta O_{D^{Tr},\mathsf{S}}(s^2_M) \in \mathbb{N}$ to denote \textbf{increase in majority $z_{w^*}$ variance} where $\Delta O_{D^{Tr},\mathsf{S}}(s^2_M)= \left\lfloor \log_{10}\left(\frac{s^2(\mathsf{S}_M)}{s^2(\mathsf{D}^{Tr}_M)}\right) \right\rfloor \in \mathbb{N}$.  

Fig. \ref{fig:zEcoli}, for example, expresses Ecoli's increase in minority $z_{w^*}$ variance of more than two orders of magnitude where $\Delta O_{D^{Tr},\mathsf{S}}
(s^2_m) = \left\lfloor \log_{10}\left(\frac{14.68}{0.04}\right) \right\rfloor = 2.1$. The concentration of minority $z_{w^*}$ values (red columns) in the top histogram around $z_{w^*} = -9$ reflects the low variance of unsampled minority training data; i.e., $s^2(D_m^{Tr}) = 0.04$. In the bottom histogram in Fig. \ref{fig:zEcoli}, the red columns spanning from approximately $z_{w^*} = 3$ to $z_{w^*}=18$ reflects higher variance of $s^2(\mathsf{S}_m) = 14.68$. 

To show MOODS's ability to improve overall feature diversity, we use the following definition for increases in $z_{w^*}$ variance from $D^{Tr}$ to $\mathsf{S}$ including both minority and majority increases.

\begin{definition}\label{Def: deltaVariance}
    Let $\Delta O(s^2) \in \mathbb{N}$ denote the \textbf{ order of magnitude increase in $z_{w^*}$ variance}: 
    \begin{align*}
        \Delta O_{D^{Tr}, \mathsf{S}}(s^2) &:=\frac{1}{2}\left(\Delta O_{D^{Tr},\mathsf{S}}(s^2_m) + \Delta O_{D^{Tr},\mathsf{S}}(s^2_M)\right).
    \end{align*}
\end{definition}

Putting it together, our method's ability to resample originally imbalanced data $D^{Tr}$ into a training set $\mathsf{S}$ that is both $\epsilon$ non-overlapping and $\delta$ diverse is measured with the following: 
\begin{definition}\label{def:setOptimal}
    Set $\mathsf{S}$ has attained \textbf{$\epsilon/ \delta$ non-overlapping diversification (${\mathsf{S}}_{\epsilon/ \delta}$)} if and only if 
    \begin{enumerate}
        \item minority $z_{w^*}$ overlap has decreased by at least $\epsilon$; i.e., $\Delta_{D^{Tr}, \mathsf{S}}(\kappa_m) \geq \epsilon$; and 
        \item $z_{w^*}$ variance has increased by at least $\delta$ orders of magnitude; i.e., $\Delta O_{D^{Tr}, \mathsf{S}}(s^2) \geq \delta$.
    \end{enumerate} 
\end{definition}

\section{Evaluation} \label{evaluationSection}
\begin{wrapfigure}{1}{0.4\textwidth}
    \centering
    \vspace{-5pt}
    \includegraphics[width = 0.37\textwidth]{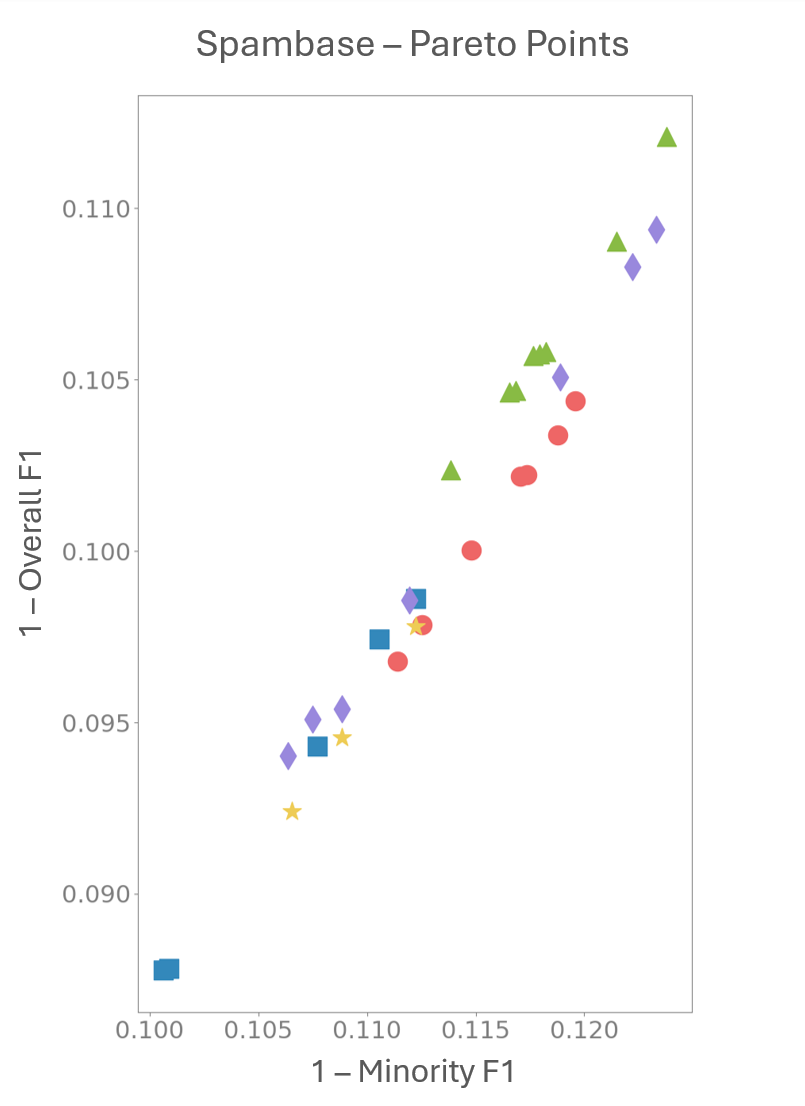}
    \caption{Pareto points of five runs of the MOODS algorithm on the Spambase dataset are represented by five color/shape combos (ex: green triangles). For each run, the initial point is located in the upper-right while the last is in the lower-left. This consistent linear descent indicates MOODS's ability to balance its two objectives, improving both minority and overall $F1$ scores.} 
    \label{fig:Pareto}
    \vspace{-5pt}
\end{wrapfigure}

We present an evaluation of our proposed method supporting our multi-objective bilevel optimization sampling approach. As shown in Table \ref{fig:moodsResults}, our method achieves results competitive with state-of-the-art sampling methods SMOTified-GAN \cite{sharma_smotified-gan_2022}, GAN-based Oversampling (GBO) \cite{ahsan_enhancing_2024}, SVM-SMOTE-GAN (SSG) \cite{ahsan_enhancing_2024}, and Majority Undersampling with Bilevel Optimization (MUBO) \cite{medlin_bilevel_2024}. Performance results are measured with $F1$ scores (Def. \ref{def: F1score}) using testing data $D^{Te}$. Additionally, with the $\epsilon/ \delta$ non-overlapping diversification metric (Def. \ref{def:setOptimal}), we demonstrate that our method's state-of-the-art performance is proportional to more diverse and less overlapping resampled training sets.

\subsection{Experimental Set-up}\label{section: setUp}
We evaluate our proposed method with seven benchmark datasets from \cite{kelly2023uci} ranging in size from 335 to 376,640 data points; in dimension from 7 to 57 features; and in imbalance ratio from 0.294\% to 39.39\% (Table \ref{fig:dataSets}). The state-of-the-art techniques to which we compare our approach - SMOTified-GAN, GBO and SSG, and MUBO - were also initially evaluated using these datasets.

The neural network in MOODS has two hidden linear layers with 256 and 128 neurons per layer; a rectified linear unit function; an additional linear layer with 128 neurons; and a softmax layer. The same model architecture is used for all methods and all datasets. The only difference is the number of neurons in the input and the output layers that vary according to the size of the dataset. For the loss function, we utilize the binary cross entropy with a data batch size of 32 and an initial learning rate of $1e-4$ using Adam optimizer \cite{kingma_adam_2017}. This set-up aligns with those found in the SMOTified-GAN \cite{sharma_smotified-gan_2022}, GBO and SSG \cite{ahsan_enhancing_2024}, and MUBO \cite{medlin_bilevel_2024}. To generate robust results, there were five runs for each dataset with a random seed generated at the beginning of each run. Run times ranged from no more than two hours for the smallest dataset to less than two days for the largest.

\begin{table}[htb]
\centering
\caption{Description of Datasets}
\label{fig:dataSets}
\begin{small}
\begin{tabular}{lrrr@{\hskip 1cm}lrrr}
\toprule
\textbf{Dataset (D)} & \textbf{|D|} & \textbf{n} & \textbf{Min. (\%)} & \textbf{Dataset (D)} & \textbf{|D|} & \textbf{n} & \textbf{Min. (\%)} \\
\midrule
Ecoli       &   335   &  7  &  5.97   & Spambase   &  4,601   & 57  & 39.39 \\
Yeast       &   513   &  8  &  9.94   & Shuttle    & 58,000   &  9  &  0.29 \\
Winequality &   655   & 11  &  2.74   & Connect4   &376,640   & 42  &  3.84 \\
Abalone     & 4,177   &  8  & 20.10   &            &          &     &       \\
\bottomrule
\end{tabular}
\end{small}
\end{table}
Developed in Python with PyTorch \cite{paszke_pytorch_2019}, code for seven methods - without sampling, SMOTE, SVM-SMOTE, GBO, SSG, MUBO, and MOODS - was successfully run on a Linux-based cluster that uses an Intel(R) Xeon(R) CPU E5-2623 v4 2.60GHz processor and a NVIDIA GeForce GTX 1080 GPU card. We were unable to execute SMOTified-GAN's experiments on our machine and instead used their published results \cite{sharma_smotified-gan_2022} in Table \ref{fig:moodsResults}. Our code is publicly available at \url{https://anonymous.4open.science/r/MOODS-EC46}.

\subsection{$F1$ Score Results}
As shown in Table \ref{fig:moodsResults}, the average $F1$ scores for our method surpass those of comparable sampling techniques on six imbalanced datasets. MOODS's $F1$ scores range from $0.62$ to $0.96$ (with variances from $2.58e{-6}$ to $2.37e{-4}$), a $1\%$ to $15\%$ improvement over the second best method. The one dataset for which our approach does not lead -- Spambase -- is, in fact, the least imbalanced with $\sim 40\%$ minority data. In comparison with other state-of-the-art, MOODS met SMOTified-GAN's and MUBO's $F1$ scores with Spambase at $0.92$ while achieving an $F1$ score within the margin of error of GBO and SSG at only $1\%$ lower. As shown in Table \ref{fig:moodsResults}, Spambase's low $F1$ scores can be accounted for by its persistent $z_{w^*}$ minority overlap percentage (Def. \ref{def:minorityOverlap}), which only modestly decreased from 13\% to 10\%, and lack of feature diversity, which increased by only 1.1 order of magnitude (Def. \ref{Def: deltaVariance}). Note also that there was no attempt made in this paper to optimize the hyper-parameters of the model. Such optimization should invariably improve $F1$ scores. However, this is not the focus of this paper. The key focus of this paper is the improvement in feature diversity and minority overlap of the training set, which is substantial for all but one of the datasets presented here.

\begin{table*}[htb]
\caption{$F1(D^{Te})$ scores compare MOODS to a technique without sampling (w/o), SVM-SMOTE (S-SMO), and four state-of-the-art methods – SMOTified-GAN (SMO-G), GBO, SSG, and MUBO. Experiments across all methods are with $20\%$ testing data. Decreases in $z_{w^*}$ minority overlap ($\Delta(\kappa_m) = \Delta_{D^{Tr},\mathsf{S}}(\kappa_m)$; Def. \ref{def:minorityOverlap}) and increases in $z_{w^*}$ variance ($\Delta O(s^2) = \Delta O_{D^{Tr},\mathsf{S}}(s^2)$; Def. \ref{Def: deltaVariance}) provide an explanation for MOODS's strong $F1$ scores.}
\label{fig:moodsResults}
\renewcommand{\arraystretch}{1.17}
\setlength{\tabcolsep}{4pt}
\centering
\begin{tabular}{>{\itshape}lcccccccccc}
    \toprule
    \textup{Dataset}  & \text{w/o} & \text{S-SMO} & \text{SMO-G} & \text{GBO} & \text{SSG} & \text{MUBO} & \textbf{MOODS} & \cellcolor{gray!10}$\Delta(\kappa_m)$ & \cellcolor{gray!10}$\Delta O(s^2)$ \\
    \midrule
    Ecoli      & 0.49 & \textit{0.77} & \textbf{0.92} & \textit{0.77} & 0.65 & 0.73 & \textbf{0.92} & \cellcolor{gray!10}100 & \cellcolor{gray!10}2.1 \\
    Yeast      & 0.47 & 0.83 & 0.82 & 0.81 & \textit{0.85} & 0.81 & \textbf{0.92} & \cellcolor{gray!10}95 & \cellcolor{gray!10}2.6 \\
    Winequality& 0.49 & \textit{0.58} & 0.53 & \textbf{0.62} & 0.48 & 0.54 & \textbf{0.62} & \cellcolor{gray!10}100 & \cellcolor{gray!10}3.1 \\
    Abalone    & 0.84 & 0.82 & 0.76 & 0.83 & 0.83 & \textit{0.86} & \textbf{0.87} & \cellcolor{gray!10}26 & \cellcolor{gray!10}2 \\
    Spambase   & \textit{0.92} & 0.91 & \textit{0.92} & \textbf{0.93} & \textbf{0.93} & \textit{0.92} & \textit{0.92} & \cellcolor{gray!10}3 & \cellcolor{gray!10}1.1 \\
    Shuttle    & 0.50 & 0.50 & - & \textit{0.88} & 0.87 & 0.60 & \textbf{0.89} & \cellcolor{gray!10}100 & \cellcolor{gray!10}1.7 \\
    Connect4   & 0.87 & 0.90 & 0.90 & 0.94 & \textit{0.95} & \textbf{0.96} & \textbf{0.96} & \cellcolor{gray!10}52 & \cellcolor{gray!10}0.7 \\
    \bottomrule
\end{tabular}
\end{table*}

In terms of minority and majority performance as depicted in Table \ref{fig:moodsMinMajResults}, our approach led with both its $F1_m$ and $F1_M$ scores on three datasets -- Ecoli, Yeast, and Connect4 -- and with its $F1_M$ score on two additional datasets -- Abalone and Shuttle. Abalone's lower $F1_m$ score can be accounted for by its final $z_{w^*}$ minority overlap percentage of 8\%. However, this minority overlap represents a significant decrease of 26\%, and Abalone's feature diversity increased by two orders of magnitude, keeping Abalone's overall $F1$ score on top. While MOODS's overall $F1$ score for the Wine Quality dataset dominates, its $F1_m$ and $F1_M$ scores fall short. Winequality's $F1_m$ score is within the margin of error compared to other methods with the exception of MUBO. MOODS's $F1_m$ score is $-45\%$ that of MUBO; however, MOODS's $F1_M$ score more than makes up for it at $+60\%$. Similarly for Winequality's $F1_M$ score, MOODS is $-5\%$ and $-3\%$ compared to w/o sampling and SSG, respectively; however, MOODS's $F1_m$ scores are $+31\%$ higher. In terms of the quality of its final training set, Winequality's is excellent with a 100\% decrease in $z_{w^*}$ minority overlap and three orders of magnitude increase in $z_{w^*}$ variance.

\begin{table*}[htb]
\caption{$F1_m(D^{Te})$ and $F1_M(D^{Te})$ scores provide a comparison of MOODS to a technique without sampling (w/o); two classic sampling techniques – SMOTE and SVM-SMOTE (S-SMO) – and three state-of-the-art methods – GBO, SSG and MUBO.}
\label{fig:moodsMinMajResults}
\centering
\begin{tabular}{>{\itshape}l lccccccc}
 \toprule
 \textup{Dataset} & \text{Metric} & \text{w/o} & SMOTE & S-SMO & GBO & SSG & MUBO & MOODS \\
 \midrule
 Ecoli      & $F1_m$ & 0.00 & 0.40 & 0.57 & 0.57 & 0.33 & 0.67 & 0.86 \\
            & $F1_M$ & 0.98 & 0.98 & 0.97 & 0.97 & 0.97 & 0.80 & 0.99 \\[4pt]
 Yeast      & $F1_m$ & 0.75 & 0.67 & 0.71 & 0.67 & 0.74 & 0.77 & 0.86 \\
            & $F1_M$ & 0.95 & 0.93 & 0.95 & 0.95 & 0.96 & 0.84 & 0.98 \\[4pt]
 Winequality& $F1_m$ & 0.00 & 0.31 & 0.25 & 0.32 & 0.00 & 0.76 & 0.31 \\
            & $F1_M$ & 0.98 & 0.89 & 0.91 & 0.92 & 0.96 & 0.33 & 0.93 \\[4pt]
 Abalone    & $F1_m$ & 0.74 & 0.65 & 0.72 & 0.73 & 0.74 & 0.87 & 0.80 \\
            & $F1_M$ & 0.94 & 0.87 & 0.92 & 0.93 & 0.92 & 0.85 & 0.94 \\[4pt]
 Spambase   & $F1_m$ & 0.89 & 0.90 & 0.89 & 0.91 & 0.92 & 0.92 & 0.90 \\
            & $F1_M$ & 0.94 & 0.92 & 0.93 & 0.95 & 0.94 & 0.92 & 0.93 \\[4pt]
 Shuttle    & $F1_m$ & 0.00 & 0.72 & 0.00 & 0.79 & 0.75 & 0.40 & 0.77 \\
            & $F1_M$ & 1.00 & 1.00 & 1.00 & 0.97 & 0.99 & 0.80 & 1.00 \\[4pt]
 Connect4   & $F1_m$ & 0.75 & 0.82 & 0.82 & 0.89 & 0.90 & 0.91 & 0.93 \\
            & $F1_M$ & 0.99 & 1.00 & 0.98 & 0.99 & 1.00 & 1.00 & 1.00 \\
 \bottomrule
\end{tabular}
\end{table*}

\begin{table*}[htb]
\renewcommand{\arraystretch}{1.17} 
\caption{Results for the $\epsilon/ \delta$ metric for non-overlapping diversification (Def. \ref{def:setOptimal}) verify MOODS's ability to deliver optimal training data. Details below include the \textbf{minority $z_{w^*}$ overlap percentage} for MOODS's final sample $\mathsf{S}$--$\kappa_m(\mathsf{S})$--and change in minority and majority $z_{w^*}$ variance -- $\Delta O_{D^{Tr},\mathsf{S}}(s^2_m)$ and $\Delta O_{D^{Tr},\mathsf{S}}(s^2_M)$, respectively. MOODS's ability to deliver balanced training was also verified below with imbalanced ratios where $\frac{|\mathsf{S}_m|}{|\mathsf{S}|} \sim 0.5$. Note that $\Delta(\kappa_m))= \Delta_{D^{Tr},\mathsf{S}}
(\kappa_m)$ (Def. \ref{def:minorityOverlap}) and $\Delta O
(s^2) = \Delta O_{D^{Tr},\mathsf{S}}
(s^2)$ (Def. \ref{Def: deltaVariance}).  }
\label{fig:optimalityResults}
\centering 
\begin{tabular}{l c c c c c }
     \toprule
     \textup{Dataset}  & $\frac{|\mathsf{S}_m|}{|\mathsf{S}|}$  & $\Delta
(\kappa_m) = \kappa_m(D^{Tr}) -\kappa_m(\mathsf{S})$ & $\Delta O
(s^2) = \frac{1}{2}\left(\Delta O(s^2_m) + \Delta O(s^2_M)\right)$ \\
     \midrule
     Ecoli & 0.48 & 100 = 100 - 0 & $2.1 = \frac{1}{2}(2.52  + 1.59)$  \\
     
     Yeast & 0.5 & 95 = 100 - 5 & $2.6 = \frac{1}{2}(2.39 + 2.89)$\\
     Winequality & 0.5 & 100 = 100 - 0 & $3.1 = \frac{1}{2}(3.24 + 3.02)$\\
     Abalone & 0.5 & 26 = 34 - 8 & $2.0 = \frac{1}{2}(2.05 + 1.95)$ \\
     Spambase & 0.5 & 10 = 13 - 3 & $1.1 = \frac{1}{2}(0.67 + 1.43)$ \\
     Shuttle & 0.46 & 100 = 100 - 0 & $1.7 = \frac{1}{2}(2.04 + 1.26)$ \\
     Connect4 & 0.5 & 51.7 = 52 - 0.3 & $0.7 = \frac{1}{2}(-.03 + 1.50)$ \\
     \bottomrule
    \end{tabular}
\end{table*}

\subsection{Non-overlapping Diversification ($\mathsf{S}_{\epsilon/ \delta}$) Results}\label{optimalResults}
Considering $z_{w}$ model outputs as 1-D analogs to input training data, the histogram plots comparing $z_{w^*}$ model outputs of initial, imbalanced data with those of MOODS-sampled data -- Fig. \ref{fig:zEcoli} (Sec. \ref{section: Intro}) and Figs. \ref{fig:zYeast} - \ref{fig:zConnect4} (Technical Appendix \ref{appendix}) -- verify MOODS's ability to overcome the fundamental problems of overlap and lack of diversity. MOODS is able to construct balanced training sets with less overlap having moved minority $z_{w^*}$ values over the $z_{w^*} = \frac{1}{2}$ decision boundary to their correct side. Additionally, the figures show how training sets resampled by MOODS have increased levels of feature diversity with the histogram columns spread across the $z_{w^*}$ axis. 

While the figures provide qualitative results, we also present quantitative results measuring MOODS's ability to construct $\mathsf{S}_{ \epsilon/ \delta}$ training sets as defined in Def. \ref{def:setOptimal} and amassed in Table \ref{fig:moodsResults} (Sec. \ref{evaluationSection}) and Table \ref{fig:optimalityResults} (Technical Appendix \ref{appendix}). We attained five $\mathsf{S}_{26/1.7}$ training sets whose $z_{w^*}$ minority overlap decreased by at least $26\%$ and $z_{w^*}$ variance increased by at least 1.7 orders of magnitude. Cut another way, MOODS attained four $\mathsf{S}_{52/0.7}$ ultimate training sets whose $z_{w^*}$ minority overlap and variance improved by at least $52\%$ and 0.7 orders of magnitude, respectively. Together these six datasets - Ecoli, Yeast, Winequality, Abalone, Shuttle, and Connect4 - span in size from 335 to 176,640 and in imbalance ratio from 0.294\% to 20.1\%. 

On the other hand, Spambase's decrease in $z_{w^*}$ minority overlap of only $3\%$ and modest increase in $z_{w^*}$ variance of $10^{1.1}$ could not be overcome. While Abalone had the next lowest decrease in minority overlap of $26\%$, its feature variance made up for it by increasing by  $10^{2}$. Spambase's $\mathsf{S}_{3/1.1}$ result accounts for its less dominant $F1$ scores. To note, Spambase's $F1$ scores are the most consistent across all classification methods with both $F1_m$ and $F1_M$ being within $3\%$ of one another. As such, Spambase is a candidate for requiring a model more complex than the one currently being used by MOODS and the other state-of-the-art methods SMOTified-GAN, GBO, SSG, and MUBO to move the needle on its $F1$ scores.

\section{Conclusion and Future work}
 
Using specially formulated multi-objective bilevel optimization problems to motivate an algorithm's selection of training data provides a promising approach to data sampling with broad applications. Framing $z_{w^*}$ model output values as 1-D proxies to $x \in \mathbb{R}^n$ feature vectors motivates future work in both sampling techniques and developing metrics more broadly applicable to machine learning. 

For example, future work includes extending MOODS's multi-objective bilevel optimization framework to the multi-class case for imbalanced data classification. Additionally, alternative convergence criteria are being explored for confirming convergence of our unique upper-level problem for finding optimal training data.

\section{Acknowledgments}
This material is based upon work supported by the U.S. Department of Energy, Office of Science, Office of Workforce Development for Teachers and Scientists, Office of Science Graduate Student Research program under contract number DE-SC0014664. This work was also supported in part by the U.S. Department of Energy, Office of Science, Office of Advanced Scientific Computing Research, Scientific Discovery through Advanced Computing (SciDAC) Program through the FASTMath Institute, RAPIDS institute, distributed resilience systems for science program and the integrated computational and data infrastructure for science discovery program under Contract No. DE-AC02-06CH11357.

\begin{small}

\bibliographystyle{plain}
\bibliography{main}

\medskip
\end{small}


\appendix

\section{Technical Appendices and Supplementary Material}\label{appendix}
Using $z_{w^*} \in \mathbb{R}$ model outputs as a proxy measurement for feature data $x \in \mathbb{R}^n$ (Sec. \ref{novelMetricSection}), Figures \ref{fig:zYeast} - \ref{fig:zConnect4} provide further evidence that MOODS sampling enables a model to separate training data into balanced minority and majority data with decreased overlap and increased feature diversity. Keeping model architecture constant, the histogram plots compare $z_{w^*}$ outputs of a model trained with imbalanced data (top plots) with $z_{w^*}$ outputs of a model trained with data resampled by MOODS (bottom plots) for the Yeast, Wine Quality, Abalone, Spambase, Shuttle and Connect4 datasets.

Figure captions connect the qualitative results of the plots with the quantitative results of the `$\epsilon / \delta$ non-overlapping diversification metric' amassed in Table \ref{fig:optimalityResults}. Connections with the $F1$ scores in Table \ref{fig:moodsResults} are also discussed. As defined in Sec. \ref{novelMetricSection} and discussed in Sec. \ref{optimalResults}, the `$\epsilon / \delta$  non-overlapping diversification metric'  includes a measurement for decreases in minority $z_{w^*}$ overlap ($\Delta(\kappa_m) = \Delta_{D^{Tr},\mathsf{S}}(\kappa_m)$; Def. \ref{def:minorityOverlap}) and increases in minority and majority $z_{w^*}$ variance by orders of magnitude ($\Delta O(s^2) = \Delta O_{D^{Tr},\mathsf{S}}
(s^2)$; Def. \ref{Def: deltaVariance}). MOODS's sampling method resulted in balanced training sets for all seven benchmark datasets in terms of numbers of minority and majority datapoints ($\frac{|\mathsf{S}_m|}{|\mathsf{S}|} \sim 0.5$ in Table \ref{fig:optimalityResults}).

\begin{figure}[H]
    \begin{center} 
    \begin{subfigure}[b]{0.66\textwidth}
        \includegraphics[width = \textwidth]{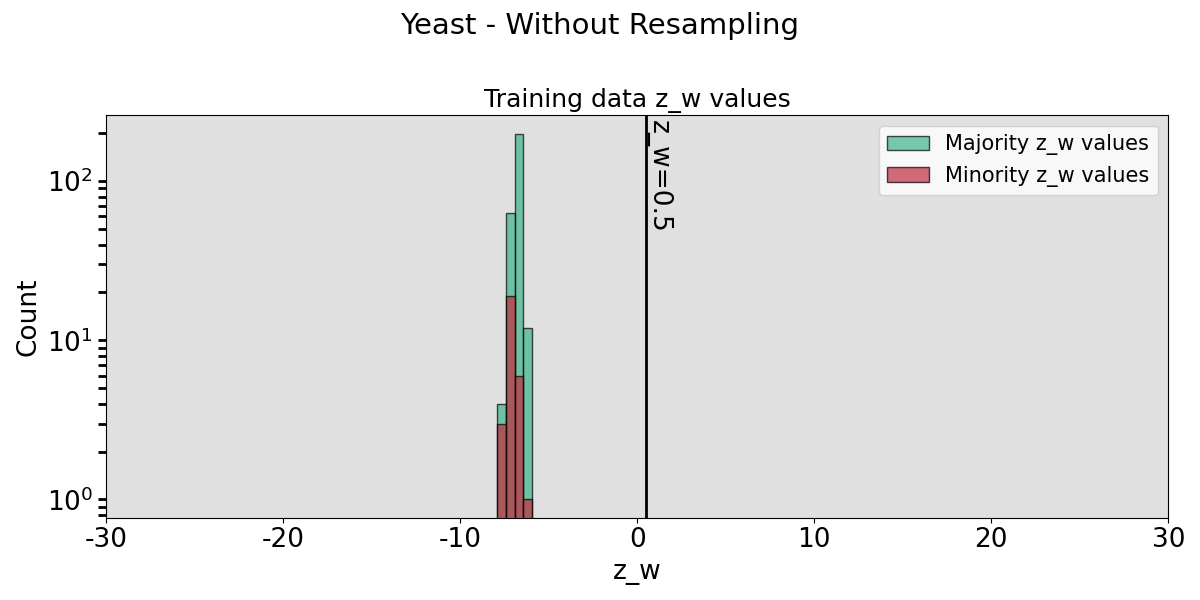}
    \end{subfigure}
    \begin{subfigure}[b]{0.66\textwidth}
        \includegraphics[width = \textwidth]{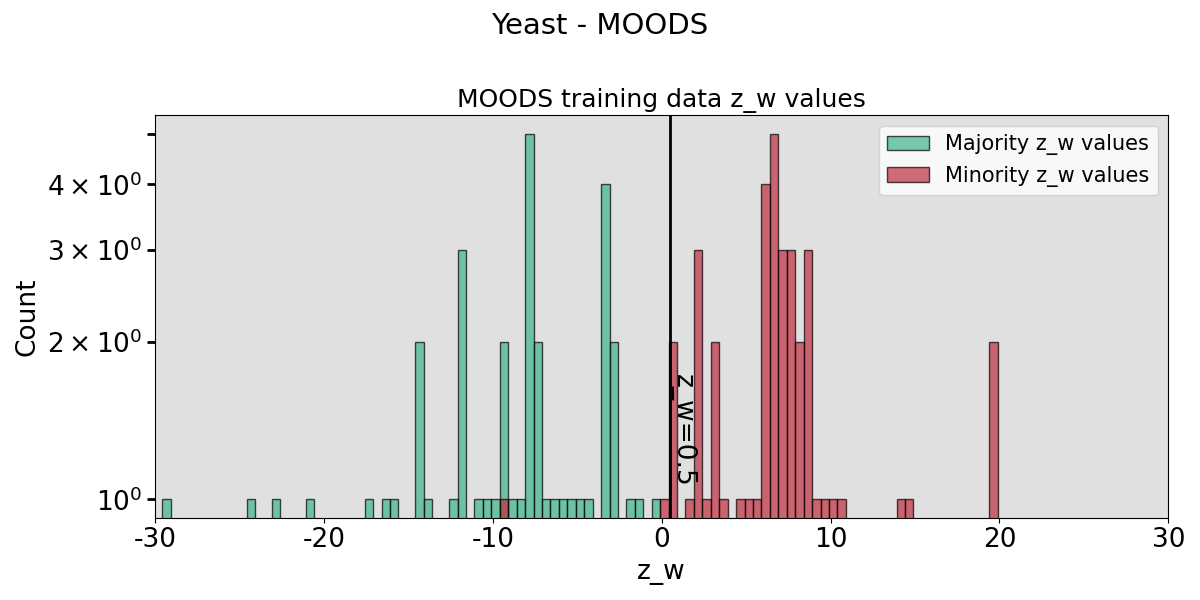}
    \end{subfigure}
    \end{center}
    \caption{Histogram plots of $z_{w^*}$ values for the Yeast dataset show how MOODS resampled training data exhibits increased feature diversity for minority (right; red) and majority (left; green) data, quantified by $z_{w^*}$ variance increase of $10^{2.6}$ ($\Delta O(s^2) = 2.6$). Increased variance and decreased minority overlap ($\kappa(S_m)=5\%$) resulted in a high $F1$ score of $0.932$.}
    \label{fig:zYeast}
\end{figure}

\begin{figure}[H]
    \begin{center} 
    \begin{subfigure}[b]{0.66\textwidth}
        \includegraphics[width = \textwidth]{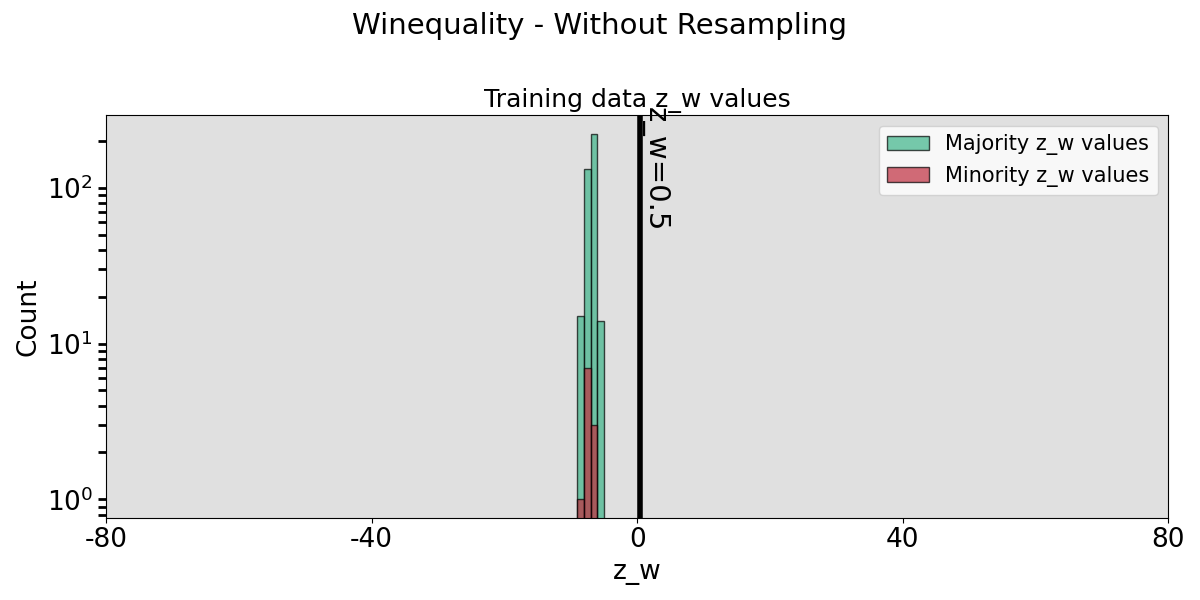}
    \end{subfigure}
    \begin{subfigure}[b]{0.66\textwidth}
        \includegraphics[width = \textwidth]{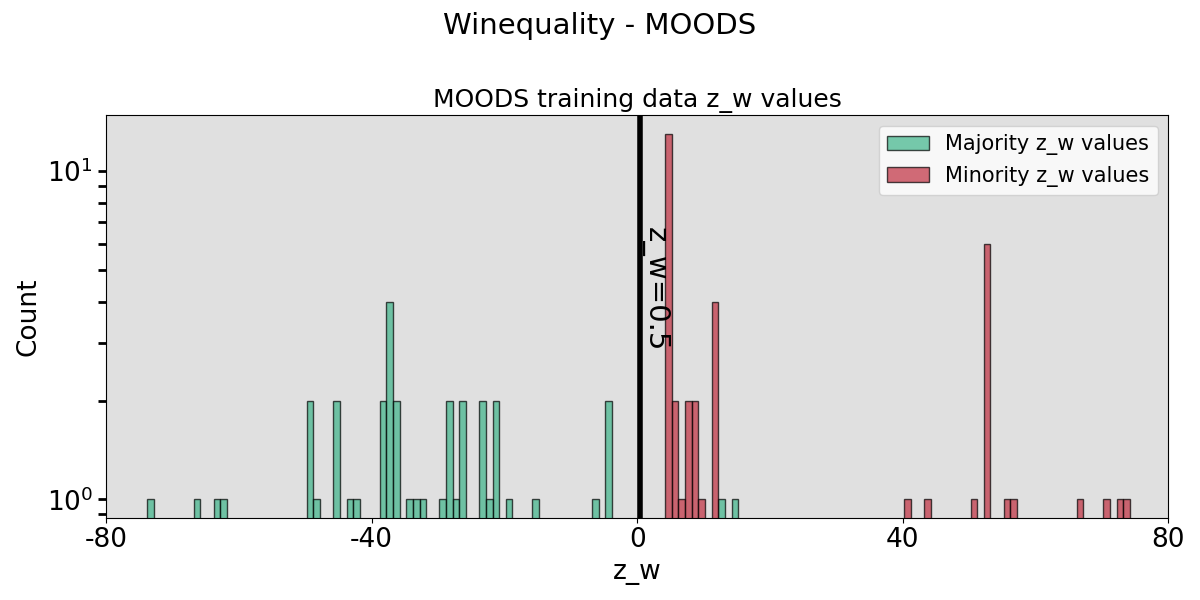}
    \end{subfigure}
    \end{center}
    \caption{Histogram plots of $z_{w^*}$ values for the Wine Quality dataset show how training data resampled by MOODS resulted in significantly increased diversity ($\Delta O(s^2) = 3.1$) and decreased minority overlap ($\kappa_m(\mathsf{S})=0\%$). Its $F1$ score of $0.624$ comes in strong compared to other methods.}
    \label{fig:zWinequality}
\end{figure}

\begin{figure}[H]
    \begin{center} 
    \begin{subfigure}[b]{0.66\textwidth}
        \includegraphics[width = \textwidth]{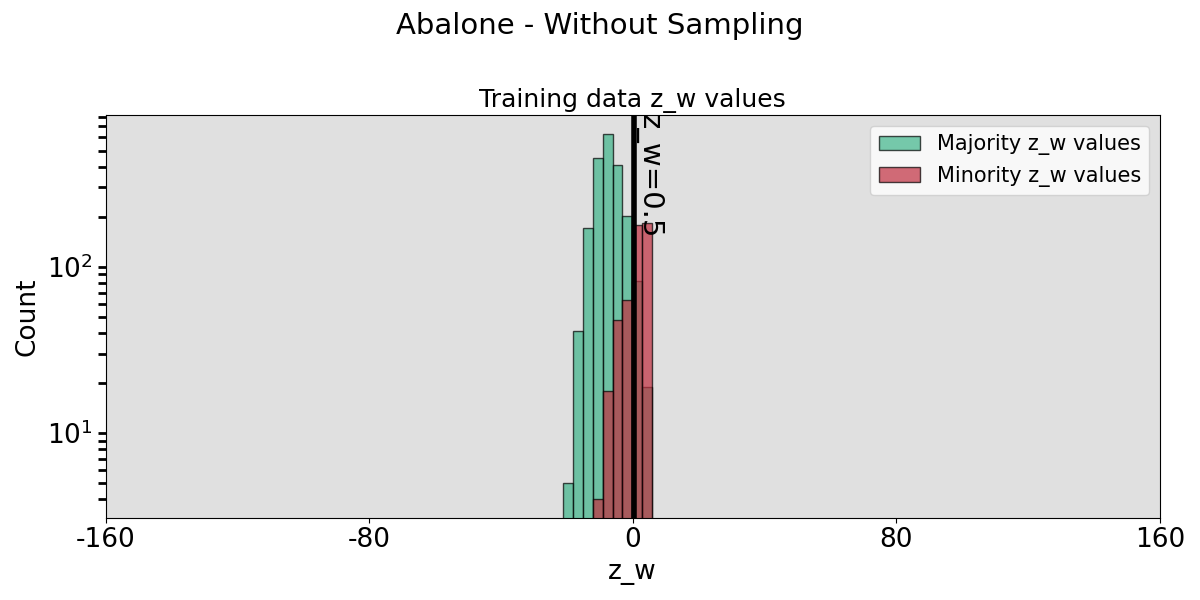}
    \end{subfigure}
    \begin{subfigure}[b]{0.66\textwidth}
        \includegraphics[width = \textwidth]{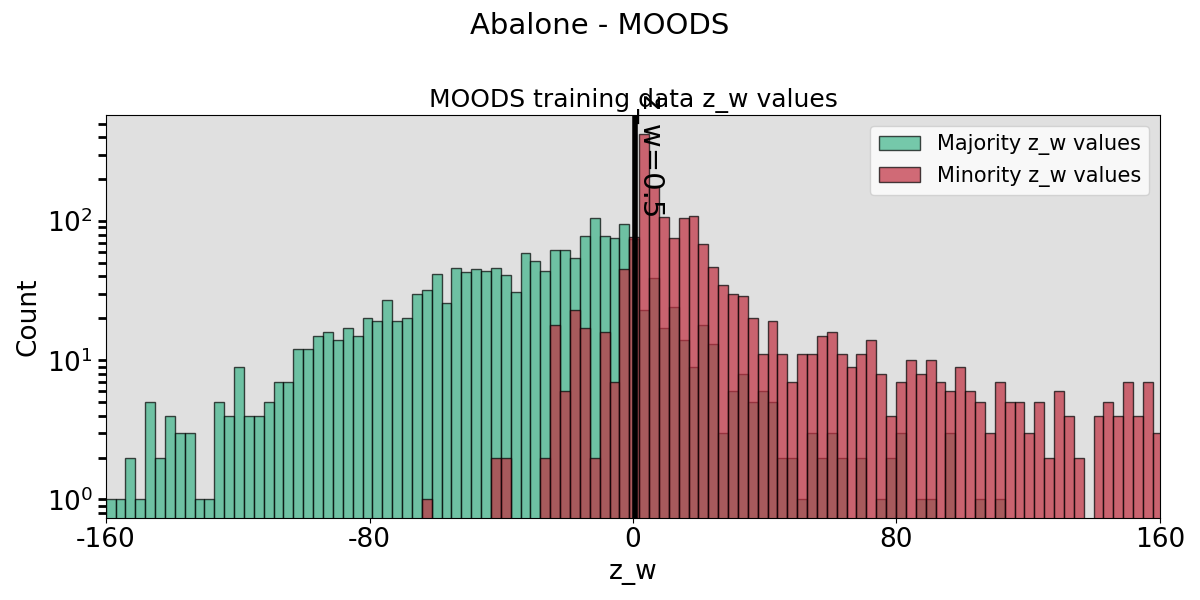}
    \end{subfigure}
    \end{center}
    \caption{Histograms of $z_{w^*}$ values for Abalone show how MOODS increased minority and majority diversity by $10^{2}$ ($\Delta O(s^2) = 2$) and decreased minority overlap of 26\%. While not eliminating overlap, the improvements in diversity and overlap resulted in a top $F1$ score of $0.87$.}
    \label{fig:zAbalone}
\end{figure}

\begin{figure}[H]
    \begin{center} 
    \begin{subfigure}[b]{0.66\textwidth}
        \includegraphics[width = \textwidth]{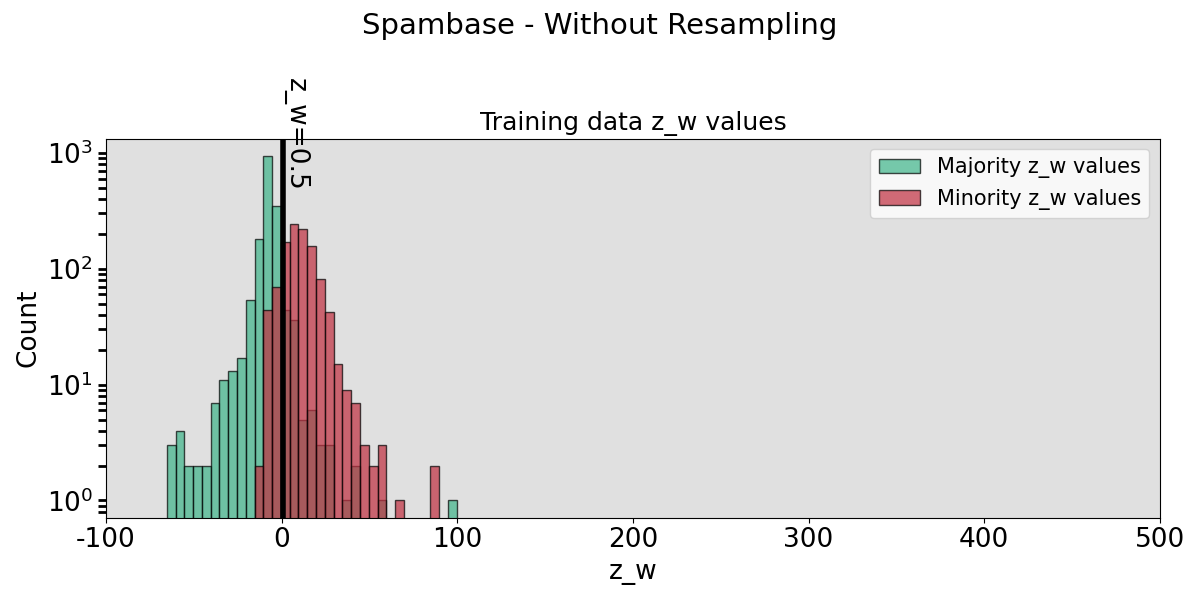}
    \end{subfigure}
    \begin{subfigure}[b]{0.66\textwidth}
        \includegraphics[width = \textwidth]{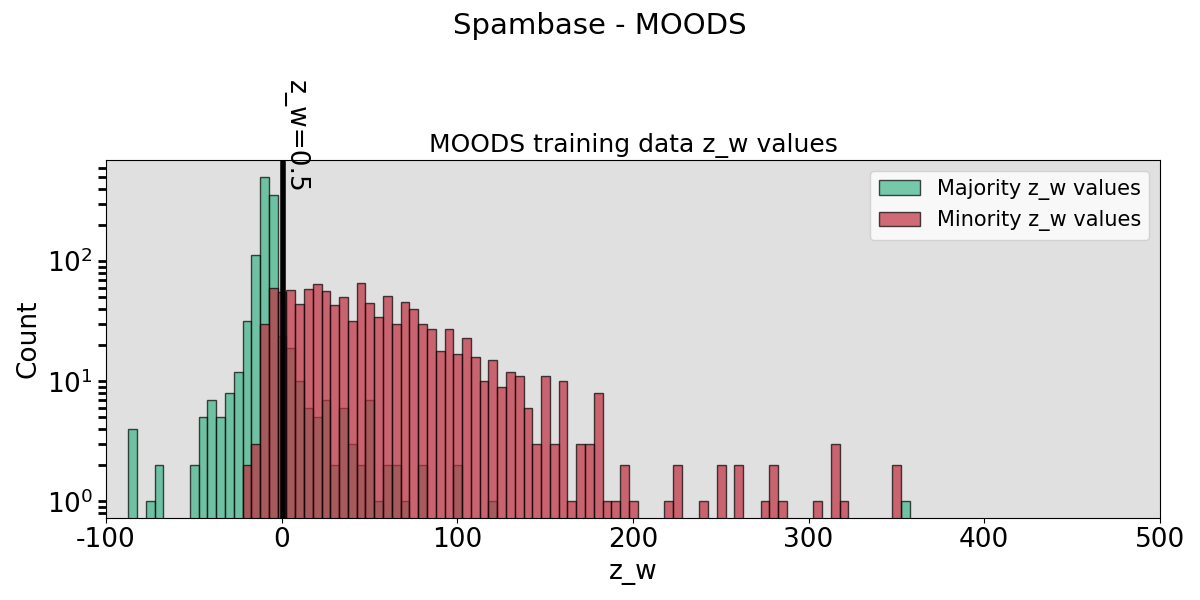}
    \end{subfigure}
    \end{center}
    \caption{Histogram plots of $z_{w^*}$ values for the Spambase dataset show how resampling training data with MOODS increased minority diversity. However, levels of change were not significant, and minority overlap persisted at $10\%$, explaining MOODS's struggle to dominate on the $F1$ chart with Spambase ($0.01$ lower than highest score) as shown in Table \ref{fig:moodsResults}.}
    \label{fig:zSpambase}
\end{figure}

\begin{figure}[H]
    \begin{center} 
    \begin{subfigure}[b]{0.66\textwidth}
        \includegraphics[width = \textwidth]{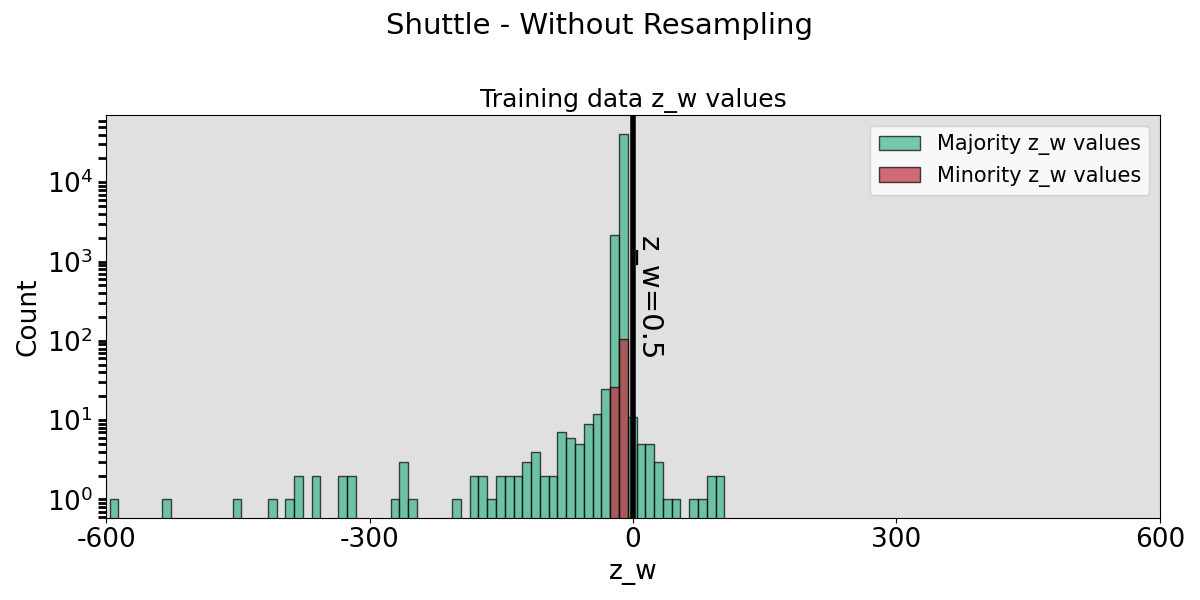}
    \end{subfigure}
    \begin{subfigure}[b]{0.66\textwidth}
        \includegraphics[width = \textwidth]{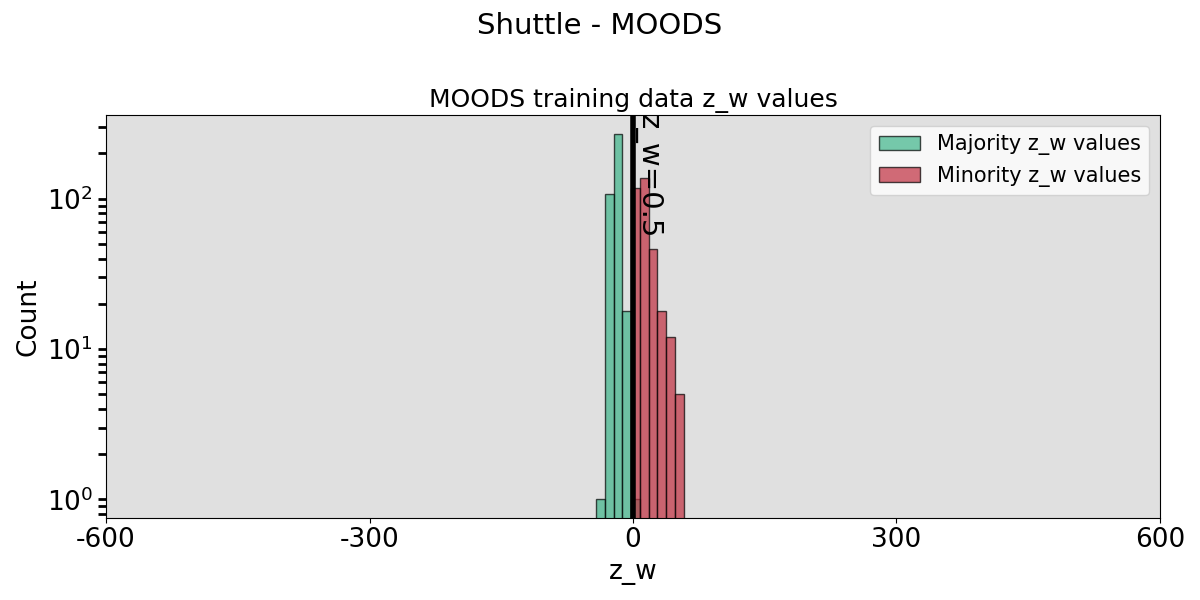}
    \end{subfigure}
    \end{center}
    \caption{Histogram plots of $z_{w^*}$ values for the Shuttle dataset show how MOODS increased minority feature diversity ($\Delta O(s_m^2) = 2.04$) and eliminated minority overlap ($\Delta (\kappa_m) = 100\%$), resulting in a leading $F1$ score of $0.89$.}
    \label{fig:zShuttle}
\end{figure}

\begin{figure}[H]
    \begin{center} 
    \begin{subfigure}[b]{0.66\textwidth}
        \includegraphics[width = \textwidth]{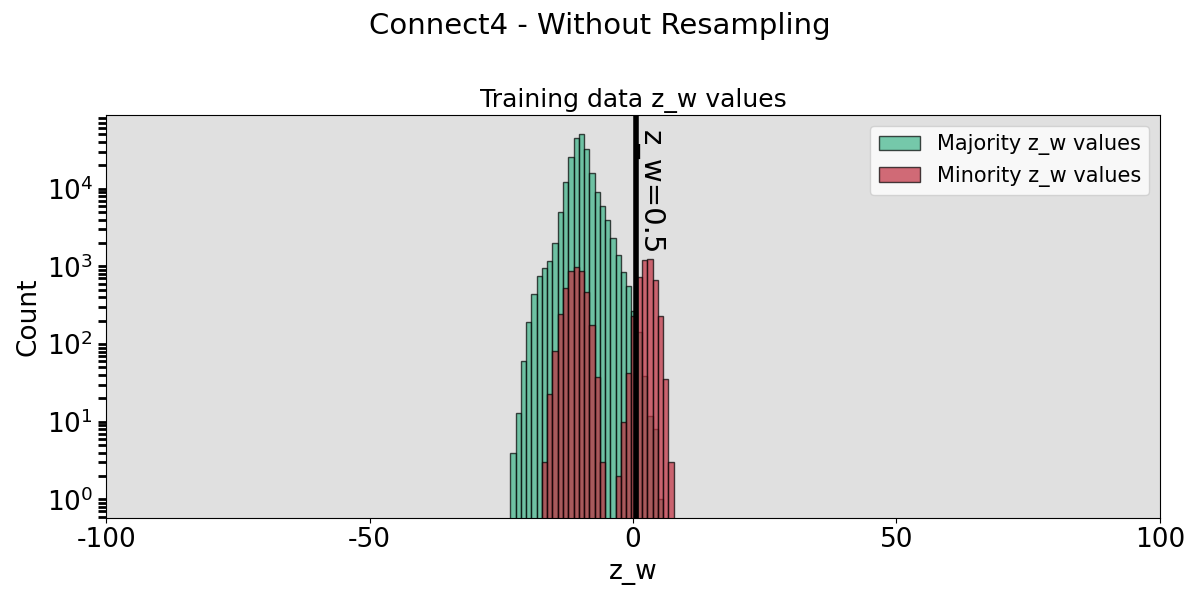}
    \end{subfigure}
    \begin{subfigure}[b]{0.66\textwidth}
        \includegraphics[width = \textwidth]{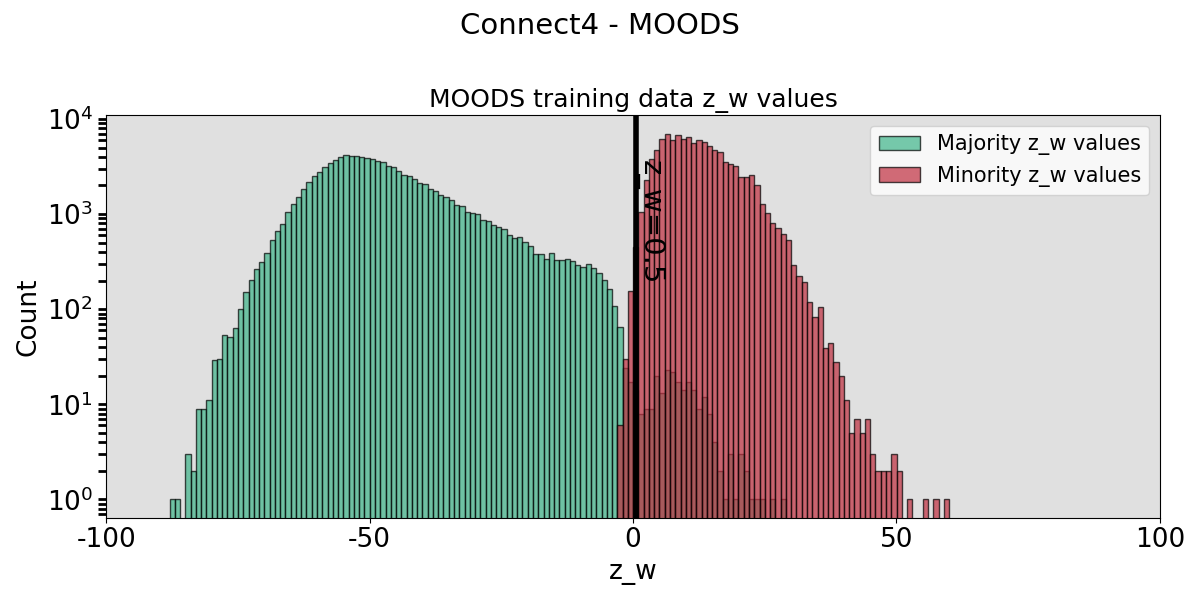}
    \end{subfigure}
    \end{center}
    \caption{Histograms of $z_{w^*}$ model outputs for Connect4 show how MOODS resampled training data (bottom plot) improved in the diversity and minority overlap of its model outputs compared with unsampled data (top plot). MOODS tied for first place with Connect4 with an $F1$ score of $0.96$.}
    \label{fig:zConnect4}
\end{figure}


\end{document}